\documentclass[letterpaper, 10 pt, conference]{ieeeconf}  

\IEEEoverridecommandlockouts                              
\usepackage[utf8]{inputenc}
\usepackage[T1]{fontenc}
\usepackage{graphicx}
\usepackage{multirow}
\usepackage{dblfloatfix}

\title{\LARGE \bf
AI-based Pilgrim Detection using Convolutional Neural Networks
}

\author{Marwa Ben Jabra $^{1}$ , Adel Ammar$^{2}$, Anis Koubaa $^{3}$, Omar Cheikhrouhou$^{4}$ , Habib Hamam$^{5}$ 
\thanks{*This work is supported by the Robotics and Internet-of-Things Lab of Prince Sultan University}
\thanks{$^{1}$ AITU American International Thelogy University in Florida, USA./ Robotics and Internet-of-Things Unit (RIoT) Labo, Saudi Arabia }
\thanks{$^{2}$ Prince Sultan University, Saudi Arabia /Gaitech Robotics,China /CISTER, INESC-TEC, ISEP, Polytechnic Institute of Porto,Portugal }
\thanks{$^{3}$Prince Sultan University, Saudi Arabia. }
\thanks{$^{4}$Taif University, Taif, Kingdom of Saudi Arabia}
\thanks{$^{5}$Faculty of Engineering, University of Moncton, CANADA}
}

\providecommand{\keywords}[1]
{
  \small	
  \textbf{\textit{Keywords---}} #1
}

\begin{document}

\maketitle
\thispagestyle{empty}
\pagestyle{empty}

\begin{abstract}
Pilgrimage represents the most important Islamic religious gathering in the world where millions of pilgrims visit the holy places of Makkah and Madinah to perform their rituals. The safety and security of pilgrims is the highest priority for the authorities. In Makkah, 5000 cameras are spread around the holy for monitoring pilgrims, but it is almost impossible to track all events by humans considering the huge number of images collected every second. To address this issue, we propose to use artificial intelligence technique based on deep learning and convolution neural networks to detect and identify Pilgrims and their features. For this purpose, we built a comprehensive dataset for the detection of pilgrims and their genders. Then, we develop two convolutional neural networks based on YOLOv3 and Faster-RCNN for the detection of Pilgrims. Experiments results show that Faster  RCNN  with  Inception  v2  feature extractor  provides  the  best  mean  average  precision  over  all classes of 51\%.

\end{abstract}

\keywords  Pilgrim Detection, Convolutional Neural Networks,
Deep Learning, You Only Look Once (Yolo), Faster R-CNN.



\section{INTRODUCTION}
Artificial Intelligence (AI) represents nowadays the hottest technology ever with a huge impact of the societies and services provided in different types of applications. One the main driving factors of artificial intelligence in the last decade is the emergence of deep learning in computer vision applications and more particularly with convolutional neural networks (CNNs). In fact, with the emergence of AlexNet \cite{alexnet} in 2012, the computer vision community aggressively moved to the application of CNN for image classification, detection, recognition and semantic segmentation. Deep learning approaches have been used in a variety of use cases namely people behavior monitoring \cite{Koubaa2019Salat}, vehicles detection \cite{benjdira2019conf, Ammar2019}, semantic segmentation of urban environments \cite{Benjdira2019Segmentation}, self-driving vehicles \cite{schoettle2014survey}, object detection and classification \cite{sevo2016, SALDANAOCHOA201953}, semantic segmentation \cite{Kampffmeyer2016, Azimi2019, Mou2018}.

In this paper, we address the problem of developing AI-based solutions for pilgrims detection and monitoring in Hajj and Umrah events, in Saudi Arabia. In fact, Hajj and Umrah attract annually millions of pilgrims from all over the world. According to Ministry of Hajj, the number of Umrah Visas issued in 2019 is around 7.5 millions and the number of pilgrims during the 5 days of the annual Pilgrimage reached 2.5 millions. The Vision 2030 of the Kingdom of Saudi Arabia aims to reach 30 millions pilgrims annually. The increasing number of pilgrims induces several challenges in terms of the security and safety of pilgrims. Although there are more than 5000 cameras spread around the holy places, it is impossible for humans to track every activity of action that would need a special intervention from security forces or from civil defense agents. There are several uses cases that would need an AI-based assistive technology to monitor pilgrims, including: (1) search and find of lost people, (2) real-time discovery of people in of emergency services, (3) assisting pilgrims in their rituals, and several others. To address this gap, we propose to develop AI-based monitoring techniques dedicated for pilgrims. We aim at the effective use of convolutional neural networks algorithms applied to video streams collected from CCTV camera of any video source containing pilgrims. The ultimate goal would be to provide an assistive technology to the authorities to promote the safety of pilgrims. 

In this paper, the contribution are three-folded. First, we built a large dataset of pilgrim and non-pilgrim instances for different genders and in different environment. Second, we have train two state-of-the-art CNN algorithms for the specific use case of pilgrim detection, namely YOLOv3 \cite{YOLOv3} and Faster R-CNN. YOLOv3 is known as begin the fastest detection algorithm, whereas Faster R-CNN \cite{Faster_R-CNN_journal} is an improvement of R-CNN \cite{R-CNN} that represents the most efficient region-based CNN algorithm for image detection. Third, we conduct a comparative study between these two algorithms to evaluate their performance in the context of pilgrim detection. 

To the best of our knowledge, this is the first paper that addresses the problem of pilgrim detection using deep learning with the state-of-the art convolutional neural networks. 

The remainder of the paper is organized as follows. Section II discusses related works on deep learning for people monitoring and existing non-AI techniques for pilgrim monitoring. Section III presents a brief background on both state of the art CNN algorithms, namely YOLOv3 and Faster R-CNN. Section IV	 presents details on the Pilgrim dataset that we built for this study. Section V presents and discussed the main results. Section VI concludes the paper and outlines future works.

\section{RELATED WORKS}

Several recent works have used CNN for people's behavior monitoring, but there were applied to contexts different from Pilgrims detection.  

Wang et al.\cite{wang2019research} were interested in the problems of the pedestrian detection and tracking failure caused by the commonly used methods of tracking. To solve this problem, for the detection, they used the Faster-RCNN framework, and for the monitoring, they used the Person-ReID method based on feature extraction and matching between different frames. This algorithm led to a tracking rate of 92.51\% on the simple standard dataset and 76.9\% on the RGB-D People dataset. 



Molchanov et al.\cite{molchanov2017pedestrian} proposed a classification approach that combines pedestrian detection and classification task in real scenes. The approach uses a YOLO neural network to overcome the problem of the low image resolution and the high density of people in a small area. 


These works present several limitations, such as (\textit{i.}) The use of high computational complexity that can be time-consuming.  To solve this problem, we use the YOLOv3, which is orders of magnitude faster. (\textit{ii.}) The low accuracy when using the RGB dataset or when dealing with a low-resolution image and the difficulty of detecting a small pedestrian. To solve this problem of detection, we used Faster R-CNN, with two different features extractor (Inception-v2 and ResNet50) that give us the best feature map that helps us to do the detection task.

On the other hand, several techniques \cite{dirgahayu2018architectural, mohandes2011pilgrims} were applied for pilgrims detections using sensing and mobile technologies, but not using deep learning methods.


Teduh et al.\cite{dirgahayu2018architectural} proposed an architecture of geo-fencing emergency alerts system for Hajj pilgrim.   The proposed architecture is based on mobile phones with GPS module, which is used as pilgrims' tracking devices. It is also created to handle the predicted load using a specific algorithm. 



Mohandes et al.\cite{mohandes2011pilgrims} developed a prototype of a wireless sensor network for tracking pilgrims in the Holy areas during Hajj. They used a principle delay tolerant network. In this system, a network of fixed master units is installed in the Holy area. Besides, every pilgrim will be given a mobile sensor unit that includes a GPS unit, a Microcontroller, antennas, and a battery that aims to sends its UID number, latitude, longitude, and time.  


These works  that were applied for pilgrims’ detections using sensing and mobile technologies also present several problems such as, (\textit{i.}) The difficulty to receive the GPS signal in some area cause problem for the pilgrim tracking system using GPS. 
(\textit{ii.}) The difficulty of working this system in a large crowd because it can’t use big data.

To solve these problems, we propose to use a computer vision deep learning for pilgrim detection in real-time. Also, it can be easily integrated to monitor pilgrims using the CCTV camera infrastructure in holy mosque areas.

\section{ALGORITHMS BACKGROUND}

For the pilgrims' detection, we are using the Faster R-CNN \cite{Faster_R-CNN_journal} and YOLOv3\cite{YOLOv3} algorithm. In this section, we present the different versions of these algorithms and the difference between them. 

\subsection{Faster R-CNN}

In this section, we provide an overview of the Faster R-CNN \cite{Faster_R-CNN_journal}  algorithm for the detection of pilgrims. It is an improved version of R-CNN \cite{R-CNN}, which has been conceived to bypass the problem of selecting a huge number of regions. This problem is inherent to the use of the conventional CNN algorithm for object detection.







\begin{figure}[h]
   \centering
   \includegraphics[width=.6\linewidth]{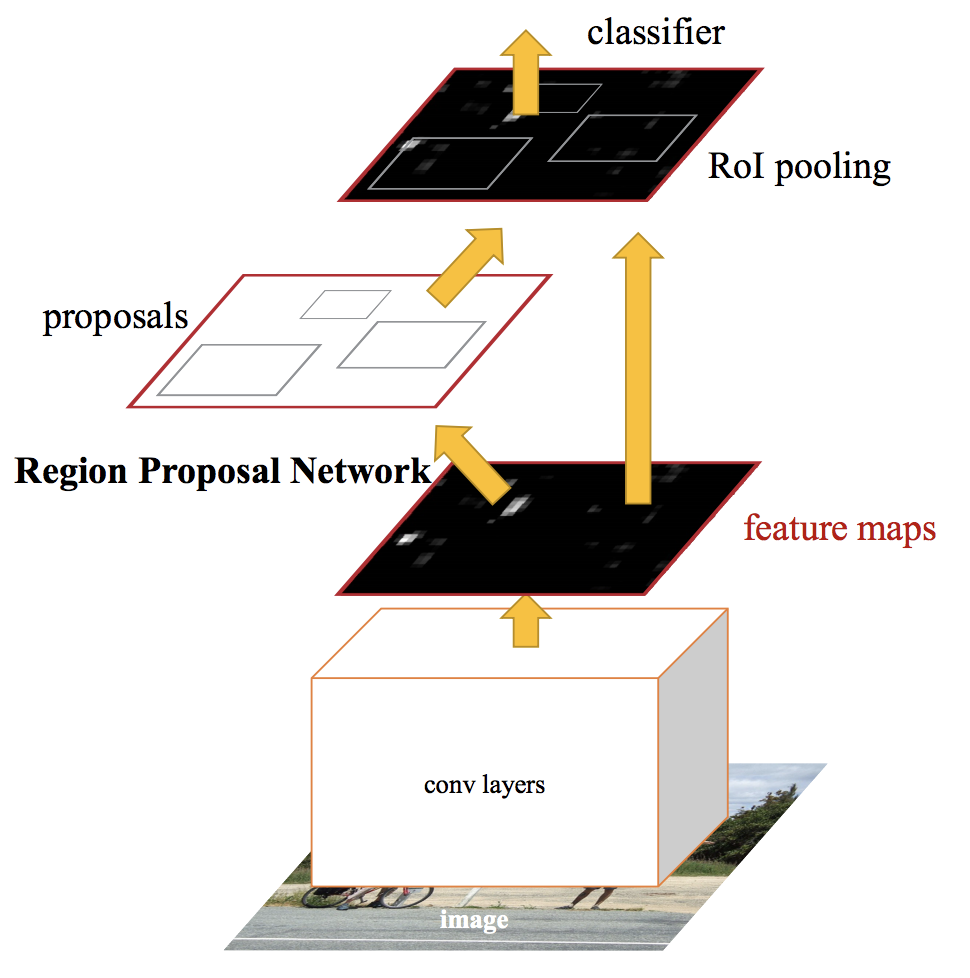}
   \caption{Faster R-CNN}
    \label{fig:Faster_R-CNN}
\end{figure}

The Faster R-CNN \cite{Faster_R-CNN_journal} algorithm presented in figure1 is the improved version of  R-CNN. This algorithm contains two modules that share the same convolutional layers. These modules are:
\begin{itemize}
\item    The region proposal network RPN
\item    A Fast R-CNN detector 
\end{itemize}

The RPN module is a fully convolutional network that aims to generate the region proposals, which are the bounding boxes that possibly include the candidate object, using multiple scales and object ratios. Each region proposal has an objectness score that measures the belonging of the region to the set of objects versus the background \cite{Ammar2019}. 
 
The Fast R-CNN detector is composed of the two following steps:

\begin{itemize}
\item    The extraction of features vectors from the region of interest ROIs using the ROI pooling.
\item    The feature vector obtained is the input of the classifier composed of fully connected layers.
\end{itemize}

The classification step output is:
\begin{itemize}
\item    A sequence of probabilities estimated of the different object considered 
\item    The coordinates of the regions proposals 

\end{itemize}

\subsection{YOLOv3}

YOLO or You Only Look Once is an improved version of convolutional neural network CNN, which is used especially for object detection, because the CNN, as originally conceived, is very time-consuming. There are three versions of YOLO. YOLOv3 \cite{YOLOv3}, which is an improved version of YOLOv2 \cite{redmon2017yolo9000} and YOLOv1 \cite{ YOLO2016}. It is characterized by:
\begin{itemize}

\item    The use of multi-label classification based on logistic regression instead of the Softmax function. 
\item    The use of cross-entropy loss function instead of the mean square error for the classification loss.
\item    The prediction of different bounding boxes based on the overlapping of the bounding box anchor with the ground truth object.
\item    The use of the concept of Feature Pyramid Network for the prediction by predicting boxes at three different scales and then extracting features from these scales. And the result of the prediction is a 3D tensor encoding the bounding box, the objectness score, and the prediction over classes.
\item    The use of Darknet-53 CNN features extractor, which is composed of 53 convolutional layers Instead of Darknet-19, using 3x3 and 1x1 filters and the skip the connection network inspired by ResNet \cite{ResNet}. 

\end{itemize}

\section{THE PILGRIMS DATASET}

In this paper, we are interested in building a comprehensive dataset for the detection of pilgrims and their genders. 


For the woman, we cannot differentiate the pilgrims from the not pilgrims because their clothes are so similar. For this purpose, we choose to put the pilgrim and not pilgrim woman in the same class.

\begin{figure}[h]
    \centering
    \includegraphics[width=.6\linewidth]{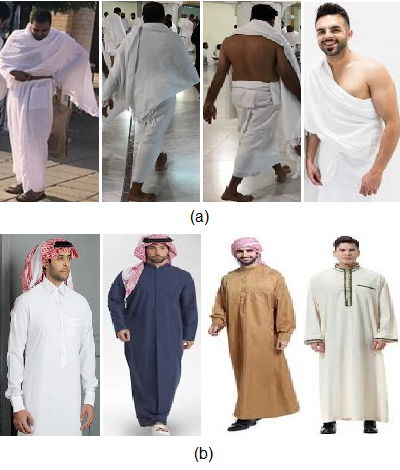}
    \caption{(a) Pilgrim Instances and (b) Non-Polgrim Instances }
    \label{fig:pilgrim}
\end{figure}

Contrariwise, the pilgrim man has specific clothes that are as different from other clothes, as we can see in Figure 3.a. For this, we choose to divide the man class into pilgrim and not-pilgrim. For the not-pilgrim class, we are focused on the white Saudi clothes, as we can see in Figure 3.b because they are quite identical, especially in term of color. 


To create our dataset, we collected 622 images of a person in the holy places of Makkah and Madinah. We choose images of persons in different environments and situations, and these images are taken from different sides and illumination.
Then, using the LabelImg software \cite{Tzutalin}, we labeled the collected dataset into three labels chosen, namely woman, pilgrim, and not-pilgrim. We obtained a dataset composed of 1165 women and 2291 man instances, which is divided into 1339 pilgrim and 952 not-pilgrim instances. The statistics of dataset instances are presented in table I.

\begin{table}[h!]
\centering
\begin{tabular}{|l|c|c|c|}
\hline
\multirow{2}{*}{} & \multirow{2}{*}{\textbf{Woman}} & \multicolumn{2}{c|}{\textbf{Man}} \\ \cline{3-4} 
 &  & \textbf{Pilgrim} & \textbf{Not pilgrim} \\ \hline
\multirow{2}{*}{\textbf{Number of instances}} & \multirow{2}{*}{1165} & \multicolumn{2}{c|}{2291} \\ \cline{3-4} 
 &  & 1339 & 952 \\ \hline
\end{tabular}
\caption{\label{tab:Table of dataset instances}Table of dataset instances}
\end{table}

Our dataset is a Pascal VOC \cite{everingham2010pascal} (Pascal object classes) dataset composed of 3 classes (woman, pilgrim, not pilgrim). We choose the Pascal VOC dataset because it enables evaluating our proposed YOLOv3 and Faster R-CNN pilgrim detection algorithm in significant variability in terms of object size, orientation, pose, illumination, position, and occlusion \cite{everingham2010pascal}. 

\section{EXPERIMENTAL EVALUATION}

In this section, we describe the results of the experimental study that we conducted to evaluate the performance of the pilgrim detection use case using two state-of-the-art algorithms, namely YOLOv3 and Faster RCNN. We start by describing the experimental setup, and we present the metrics used for the evaluation of the proposed algorithm. Finally, we analyze the results obtained for each algorithm to compare their performances \subsection{Experimental Setup}
In this experimental study, the training was done on two machines. The configurations of these two machines are presented in Table II.

\begin{table}[h]
\centering
\begin{tabular}{|l|l|l|}
\hline
\multicolumn{1}{|c|}{\textbf{}} & \multicolumn{1}{c|}{\textbf{Machine 1}} & \multicolumn{1}{c|}{\textbf{Machine 2}} \\ \hline
\textbf{CPU} & \begin{tabular}[c]{@{}l@{}}Intel Core i7-8700K \\ (3.7 GHz)\end{tabular} & \begin{tabular}[c]{@{}l@{}}Intel Core i9-9900K \\ (Octa-core)\end{tabular} \\ \hline
\textbf{Graphics card} & \begin{tabular}[c]{@{}l@{}}NVIDIA \\ GeForce 1080\\  (8 GB) GPU\end{tabular} & \begin{tabular}[c]{@{}l@{}}NVIDIA\\  GeForce RTx2080T \\ (11 GB) Gaming Cpu\end{tabular} \\ \hline
\textbf{RAM} & 32GB & 64GB \\ \hline
\textbf{Operating system} & \begin{tabular}[c]{@{}l@{}}Linux \\ (Ubuntu 16.04 TLS)\end{tabular} & \begin{tabular}[c]{@{}l@{}}Linux \\ Ubuntu 16.04 TLS)\end{tabular} \\ \hline
\end{tabular}
\caption{\label{tab: Configuration table }Configuration table}
\end{table}

For the Faster R-CNN, we choose to test two different CNN architectures for the feature extraction that is Inception-v2  \cite{ioffe2015batch} and ResNet50 \cite{ResNet}, because these are the best feature extractors for the detection task \cite{huang2017speed}. For YOLOv3, we chose to evaluate it with different resolutions, which has an impact on the accuracy and the speed of the system. We chose to use three different input sizes that have values of (320x320, 416x416, and 608x608). These settings result in five classifiers trained and tested on our pilgrim dataset. The training of these two algorithms is made to detect and recognize three classes of persons that are (\texttt{Woman}, \texttt{Pilgrim}, and \texttt{Not-Pilgrim}).
To optimize these two algorithms, we used Stochastic Gradient Descent (SGD) with a default value of momentum (0.9). 
For the learning rate, we used an initial rate of 0.001 for YOLOv3, and for the Faster R-CNN, we used an initial rate of 0.0002 with Inception-v2 and 0.0003 with ResNet50, which are the default value of each feature extractor network.
We used the weight decay value of 0.0005.

\subsection{Performance evaluation and metrics} 
For the evaluation of our proposed algorithms, we have used six metrics based on the following parameters:
\begin{itemize}

\item    \textbf{True Positive (TP)}: it is the number of instances (woman, pilgrim, and not-pilgrim) successfully detected and classified.
\item    \textbf{False Positive (FP)}: it refers to the number of instances that are wrongly classified.  
\item    \textbf{False Negative (FP)}: It is the number of non-detected instances. 

\end{itemize}
The six metrics used for the evaluation are:
\begin{itemize}
\item       $Precision=TP/(TP+FP)$
\item          $Recall=TP/(TP+FN)$
\item    $F1Score=2*Precision*Recall/(Precision+Recall)$
\item    $Quality= TP/(TP+FP+FN)$
\item    \textbf{mIoU:} mean of the  Intersection over Union that measures the overlap between the predicted and the ground-truth bounding boxes.
\item    \textbf{mAP:} mean Average Precision. Or AP (Average Precision) when it is measured on one class.  It is an approximation of the area under the precision-recall curve \cite{Ammar2019}.
\item    \textbf{FPS:} frame per second. It presents the inference speed of the algorithm.

\end{itemize}

\begin{table*}[!t]
\centering 
\begin{tabular}{|c|c|c|c|c|c|c|}
\hline
\textbf{Algorithm} & \textbf{} & \textbf{\begin{tabular}[c]{@{}c@{}}YOLOv3 \\ ( 320x320)\end{tabular}} & \textbf{\begin{tabular}[c]{@{}c@{}}YOLOv3\\ ( 416x416)\end{tabular}} & \textbf{\begin{tabular}[c]{@{}c@{}}YOLOv3\\ ( 608x608)\end{tabular}} & \textbf{\begin{tabular}[c]{@{}c@{}}Faster R-CNN\\ (Inception v2)\end{tabular}} & \textbf{\begin{tabular}[c]{@{}c@{}}Faster R-CNN\\ ( ResNet 50)\end{tabular}} \\ \hline
\multirow{9}{*}{\textbf{Class "Pilgrim"}} & FP & 20 & 33 & 27 & 19 & 24 \\ \cline{2-7} 
 & TP & 64 & 61 & 68 & 55 & 48 \\ \cline{2-7} 
 & FN & 47 & 50 & 43 & 56 & 63 \\ \cline{2-7} 
 & Precision & 0.7619 & 0.6489 & 0.7157 & 0.7432 & 0.6666 \\ \cline{2-7} 
 & Recall & 0.5765 & 0.5495 & 0.6126 & 0.4954 & 0.4324 \\ \cline{2-7} 
 & Quality & 0.4885 & 0.4236 & 0.4927 & 0.4230 & 0.3555 \\ \cline{2-7} 
 & F1score & 0.6564 & 0.5951 & 0.6601 & 0.5945 & 0.5245 \\ \cline{2-7} 
 & AP & 0.5098 & 0.4788 & 0.5398 & 0.4462 & 0.3751 \\ \cline{2-7} 
 & mIoU & 0.6352 & 0.5988 & 0.6192 & 0.5710 & 0.5850 \\ \hline
\multirow{9}{*}{\textbf{Class "Non-Pilgrim"}} & FP & 6 & 16 & 14 & 71 & 42 \\ \cline{2-7} 
 & TP & 50 & 51 & 55 & 76 & 61 \\ \cline{2-7} 
 & FN & 61 & 60 & 56 & 35 & 50 \\ \cline{2-7} 
 & Precision & 0.8928 & 0.7611 & 0.7971 & 0.5170 & 0.5922 \\ \cline{2-7} 
 & Recall & 0.4504 & 0.4594 & 0.4954 & 0.6846 & 0.5495 \\ \cline{2-7} 
 & Quality & 0.4273 & 0.4015 & 0.44 & 0.4175 & 0.3986 \\ \cline{2-7} 
 & F1score & 0.5988 & 0.5730 & 0.6111 & 0.5891 & 0.5700 \\ \cline{2-7} 
 & AP & 0.4407 & 0.4373 & 0.4786 & 0.5985 & 0.4770 \\ \cline{2-7} 
 & mIoU & 0.6352 & 0.5988 & 0.6192 & 0.5710 & 0.5850 \\ \hline
\multirow{9}{*}{\textbf{Class "Woman"}} & FP & 14 & 17 & 10 & 74 & 71 \\ \cline{2-7} 
 & TP & 45 & 59 & 42 & 97 & 86 \\ \cline{2-7} 
 & FN & 117 & 103 & 120 & 65 & 76 \\ \cline{2-7} 
 & Precision & 0.7627 & 0.7763 & 0.8076 & 0.5672 & 0.5477 \\ \cline{2-7} 
 & Recall & 0.2777 & 0.3641 & 0.2592 & 0.5987 & 0.5308 \\ \cline{2-7} 
 & Quality & 0.2556 & 0.3296 & 0.2441 & 0.4110 & 0.3690 \\ \cline{2-7} 
 & F1score & 0.4072 & 0.4957 & 0.3925 & 0.5825 & 0.5391 \\ \cline{2-7} 
 & AP & 0.2493 & 0.3295 & 0.2458 & 0.5041 & 0.4428 \\ \cline{2-7} 
 & mIoU & 0.6352 & 0.5988 & 0.6192 & 0.5710 & 0.5850 \\ \hline
\end{tabular}
\caption{\label{tab:Evaluation metrics of FASTER R-CNN and YOLOV3 for each class }Evaluation metrics of FASTER R-CNN and YOLOV3 for each class}
\end{table*}

\subsection{Comparison between Faster R-CNN and YOLO v3}

For the evaluation of the proposed algorithms, we compared the values of the six metrics for each algorithm shown in Table III and Table IV.

\begin{table*}[!t]
\centering
\begin{tabular}{|l|c|c|c|c|c|}
\hline
Algorithm & \begin{tabular}[c]{@{}c@{}}YOLOv3 \\ (320x320)\end{tabular} & \begin{tabular}[c]{@{}c@{}}YOLOv3\\ (416x416)\end{tabular} & \begin{tabular}[c]{@{}c@{}}YOLOv3\\ (608x608)\end{tabular} & \begin{tabular}[c]{@{}c@{}}Faster R-CNN\\ (Inception v2)\end{tabular} & \begin{tabular}[c]{@{}c@{}}Faster R-CNN\\ (ResNet 50)\end{tabular} \\ \hline
FP & 40 & 66 & 51 & 164 & 137 \\ \hline
TP & 159 & 171 & 165 & 228 & 195 \\ \hline
FN & 225 & 213 & 219 & 156 & 189 \\ \hline
Precision & 0.8058 & 0.7288 & 0.7735 & 0.6091 & 0.6022 \\ \hline
Recall & 0.4349 & 0.4577 & 0.4557 & 0.5929 & 0.5042 \\ \hline
Quality & 0.3905 & 0.3849 & 0.3923 & 0.4172 & 0.3744 \\ \hline
F1score & 0.5541 & 0.5546 & 0.5546 & 0.5887 & 0.5446 \\ \hline
mAP & 0.3999 & 0.4152 & 0.4214 & 0.5162 & 0.4317 \\ \hline
mIoU & 0.6352 & 0.5988 & 0.6192 & 0.5710 & 0.5850 \\ \hline
FPS & 91.28 & 65.31 & 43.84 & 3.35 & 3.8 \\ \hline
\end{tabular}
\caption{\label{tab:Evaluation metrics of FASTER R-CNN and YOLOV3 over classes}Evaluation metrics of FASTER R-CNN and YOLOV3 over classes}
\end{table*}

\subsubsection{FN, TP and FP}

Figure 3 shows that when we used the YOLOv3, the number of false negatives is much higher than the number of false positives on over classes, and also much higher than the number of true positives, which indicates that most instances go undetected.
And when using the Faster R-CNN, the number of true positives is much higher than the number of false positives and the number of false negatives on over classes, which indicates that most instances go detected.

\begin{figure} [h!]
    \centering
    \includegraphics[width=.9\linewidth]{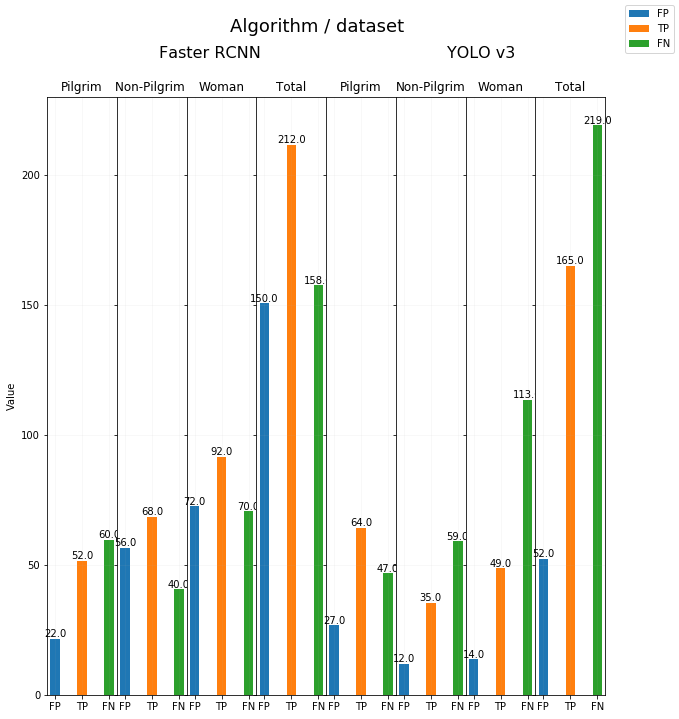}
    \caption{Average number of false positives (FP), false negatives (FN), and true positives (TP) for YOLOv3 and Faster R-CNN}
    \label{fig:FP_TP_and_FN}
\end{figure}

\subsubsection{Average Precision}

When analyzing the results, it appears that YOLOv3, with an input size of 608x608, gave a  better mAP for the Pilgrim Class and Faster R-CNN with Inception-v2 gave a better mAP on Non-Pilgrim Class (Figure 4). Figure 4 shows also that Faster R-CNN with Inception-v2 gave a much better mAP over classes.

\begin{figure} [h!]
    \centering
    \includegraphics[width=.8\linewidth]{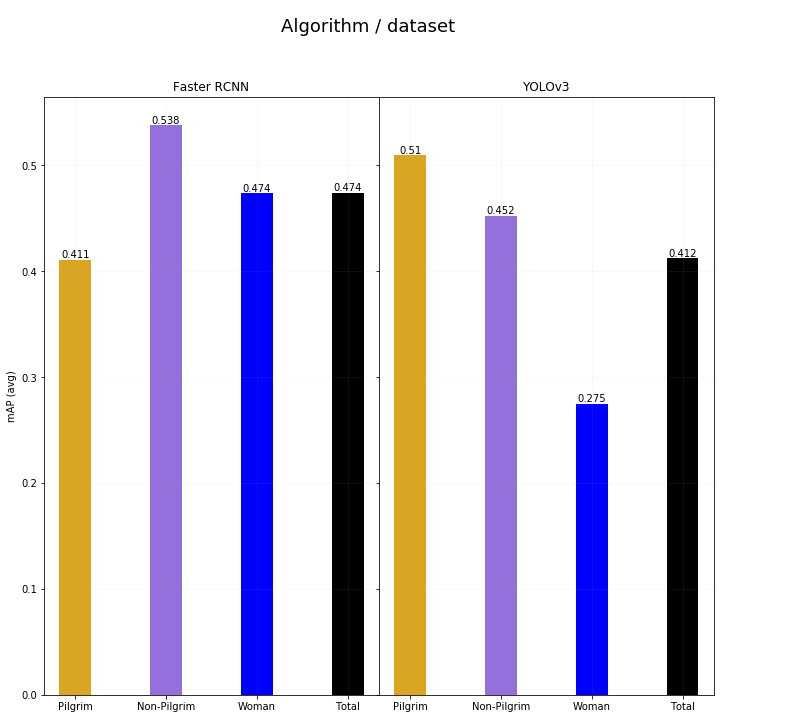}
    \caption{Comparison of the average AP between YOLOv3 ( Input size of 608x608 ) and Faster R-CNN ( Inception-v2 features extractor )}
    \label{fig:Average_AP}
\end{figure}


\subsubsection{Precision and mIoU}

The results of Average IoU, show that YOLOv3 gave a  better IoU over classes than Faster R-CNN. 
And the results of precision show that YOLOv3, with an input size of 320x320, gave a better precision for the Non-Pilgrim Class and  Faster R-CNN with Inception-v2 gave a better precision on  Pilgrim  Class. It also shows that YOLOv3, with an input size of 320x320, gave a much better precision over classes with a ratio of 80.58\%.


\subsubsection{Recall}
Analyzing the average recall results, we found that Faster R-CNN outperforms YOLOv3 in this metric with a slightly better performance with the ratio of 59.29\% for Inception-v2 feature extractor over Resnet50, and a marked inferior performance for YOLOv3 with an input size of 320x320.




\subsubsection{Robustness}

When analyzing the quality that measures the robustness of the algorithms, it appears that YOLOv3  gave a  better quality for the Non-Pilgrim Class, and Faster R-CNN gave a better Precision on Pilgrim Class. It also seems that Faster R-CNN with Inception-v2 gave a much better precision over classes with a ratio of 41.72\%.



The F1score that also measures the robustness based on the precision and the recall ratios reveals that YOLOv3, with an input size of 608x608, gave a  better performance with a ratio of 66.01\% for the Pilgrim Class and Faster R-CNN gave a better precision also on Pilgrim Class with a ratio of 59.45\%. And over all classes, Faster R-CNN with Inception-v2 gave a much better score with a ratio of 58.87\%.



\subsubsection{Inference Processing time}

The results of the average Inference speed measured in Frames per Second (FPS), for each of the tested algorithms, show that YOLOv3 is 19 times faster than Faster R-CNN in the inference phase.


\subsubsection{Effect of the feature extractor}

When analyzing the effect of the feature extractor for Faster R-CNN, it appears that  Resnet50 feature extractor is slightly faster than Inception-v2 because it is less computationally complex. But, Inception-v2 outperforms Resnet50 on almost all metrics. 

\subsubsection{Effect of the input size}
Table IV shows a significant gain in YOLOv3’s AP when moving from a 320x320 input size to 608x608. But it shows a substantial loss in YOLOv3’s precision when moving from a 320x320 input size to 608x608. That also indicates that the input size has an important impact on the inference processing speed of YOLOv3 because a larger input size generates a higher number of network parameters and operations (FPS from 43 FPS for 608*608 up to 91 FPS for 320*320).

In this section, we compared the performance of YOLOv3 (with three different input sizes) and Faster R-CNN (with two different feature extractors) and the impact of the input size and the features extractor. Figure 5 summarizes the main results of this comparison study. It compares the trade-off between AP and inference time for YOLOv3 (with three different input sizes) and Faster R-CNN (with two different feature extractors).  It can be observed that YOLOv3 (with input size 320*320) gave the best inference speed with low AP, contrary to Faster R-CNN (with Inceptionv2 as feature extractor) which gave the lowest inference speed with the best AP. This emphasizes that neither algorithm surpasses the other in all cases.

\begin{figure} [h!]
    \centering
    \includegraphics[width=.9\linewidth]{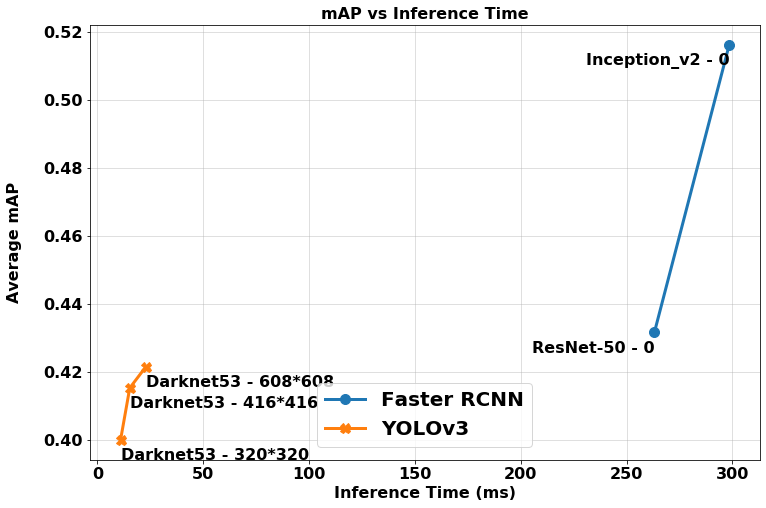}
    \caption{Comparison of the trade-off between AP and inference time for YOLOv3 (with 3 different imput sizes) and Faster R-CNN (with two different feature extractors),}
    \label{fig:mAP_vs_Inference_Time}
\end{figure}

\section{CONCLUSIONS}
In this paper, we developed convolutional neural network models for pilgrim detection for AlHajj based on YOLOv3 and Faster RCNN. We have built a dataset containing three classes of a pilgrim, non-pilgrim and women. Experimental results show that Faster RCNN with Inception v2 feature extractor provides the best mean average precision over all classes of 51\%. In our future work, we will extend the dataset to have several tens of thousands of instances to improve the overall accuracy and precision, and we will consider more classes. We also aim at developing a search application for lost people during Hajj and Umrah based on some predefined features.

\addtolength{\textheight}{-12cm}   



\section*{APPENDIX}

Appendixes should appear before the acknowledgment.

\section*{ACKNOWLEDGMENT}
This work is supported by the Robotics and Internet-of-Things Lab of Prince Sultan University.

\bibliographystyle{ieeetr}
\bibliography{ref}

\end{document}